%% file: _ecml_paper.tex
\documentclass[runningheads]{llncs}
\usepackage[T1]{fontenc}
\usepackage{graphicx}
\usepackage{booktabs}
\usepackage[misc]{ifsym}

\usepackage{mwe}

\usepackage{my_commands}
\usepackage[square,numbers]{natbib}
\usepackage{hyperref}
\usepackage{url}
\usepackage{multirow}
\usepackage{subcaption}
\usepackage{wrapfig}
\usepackage{tabularx}
\usepackage{booktabs}
\hypersetup{
    colorlinks=true,  %
    citecolor=blue, 
    linkcolor=blue,
    filecolor=blue,
    urlcolor=blue,
}

\usepackage{xargs}
\usepackage[colorinlistoftodos,prependcaption,textsize=tiny]{todonotes}
\newcommandx{\pascal}[2][1=]{\todo[linecolor=orange,backgroundcolor=orange!25,bordercolor=orange,#1]{[pascal]: #2}}
\newcommandx{\juan}[2][1=]{\todo[linecolor=red,backgroundcolor=red!25,bordercolor=red,#1]{[juan]: #2}}
\newcommandx{\nikos}[2][1=]{\todo[linecolor=blue,backgroundcolor=blue!25,bordercolor=blue,#1]{[nikos]: #2}}
\newcommandx{\ke}[2][1=]{\todo[linecolor=green!30!black!20,backgroundcolor=green!25!black!20,bordercolor=green!30!black!20,#1]{[ke]: #2}}

\renewcommand{\pascal}[2][1=]{}
\renewcommand{\juan}[2][1=]{}
\renewcommand{\nikos}[2][1=]{}
\renewcommand{\ke}[2][1=]{}

\newcommandx{\thiswillnotshow}[2][1=]{\todo[disable,#1]{#2}}

\begin{document}

\title{Single-Input Multi-Output Model Merging: Leveraging Foundation Models for Dense Multi-Task Learning}

\titlerunning{Single-Input Multi-Output Model Merging}

\author{Juan Garcia Giraldo \and
Nikolaos Dimitriadis \and
Ke Wang \and 
Pascal Frossard}

\institute{EPFL, Switzerland\\ \email{\{juan.garciagiraldo, nikolaos.dimitriadis, k.wang, pascal.frossard\}@epfl.ch}
}

\maketitle              %
\begin{abstract}
Model merging is a flexible and computationally tractable approach to merge single-task checkpoints into a multi-task model. Prior work has solely focused on constrained multi-task settings where there is a one-to-one mapping between a sample and a task, overlooking the paradigm where multiple tasks may operate on the same sample, e.g., scene understanding. In this paper, we focus on the multi-task setting with single-input-multiple-outputs (SIMO) and show that it qualitatively differs from the single-input-single-output model merging settings studied in the literature due to the existence of task-specific decoders and diverse loss objectives. We identify that existing model merging methods lead to significant performance degradation, primarily due to representation misalignment between the merged encoder and task-specific decoders. We propose two simple and efficient fixes for the SIMO setting to re-align the feature representation after merging. Compared to joint fine-tuning, our approach is computationally effective and flexible, and sheds light into identifying task relationships in an offline manner. Experiments on NYUv2, Cityscapes, and a subset of the Taskonomy dataset demonstrate: (1) task arithmetic suffices to enable multi-task capabilities; however, the representations generated by the merged encoder has to be re-aligned with the task-specific heads; (2) the proposed architecture rivals traditional multi-task learning in performance but requires fewer samples and training steps by leveraging the existence of task-specific models.
\looseness=-1
\keywords{Model Merging  \and Multi-task Learning \and Dense Prediction}
\end{abstract}

\input{main}

\end{document}

%% file: main.tex
\renewcommand{\paragraph}[1]{\subsubsection{#1}}

\section{Introduction}
\label{sec:intro}

Multi-task Learning (MTL), i.e., designing machine learning models capable of addressing multiple tasks concurrently, has gained significant attention, driven by the promise of shared representations and the practical advantages of reduced memory and inference costs \citep{Ruder_2017, Crawshaw_2020}. Dense prediction tasks, such as semantic segmentation, depth estimation, and surface normal prediction, exemplify this need, particularly in applications like autonomous driving and indoor scene understanding, where efficiency and scalability are crucial \citep{vandenhende2021multi}. However, designing multi-task systems is challenging due to the need to predefine task combinations \citep{Standley_Zamir_Chen_etal_2020,Fifty_Amid_Zhao_etal_2021}, limiting flexibility, and the prevalence of task conflicts, where optimizing one task hinders others \citep{yu2020gradient, Liu_Li_Kuang_etal_2020}.
\looseness=-1

Foundation models \citep{bommasani2021opportunities} have emerged as a compelling solution to these MTL challenges, offering general-purpose feature representations that can be adapted to a wide range of tasks via fine-tuning. 
Recent advances in model merging \citep{yadav2023ties,wang2024localizing} —combining knowledge from diverse task-specific checkpoints into a single model—have enabled multi-task capabilities without joint training, as seen in task arithmetic \citep{ilharco2023task}, which leverages linear mode connectivity \citep{frankle2020linear} to interpolate between independently fine-tuned models. These approaches have demonstrated notable success in structured and homogeneous settings, such as image classification with CLIP Vision Transformers (ViTs) \citep{radford2021learningtransferablevisualmodels,dosovitskiy2021imageworth16x16words}  or language modeling using LLaMa \citep{Touvron_Martin_Stone_etal_2023} and T5 \citep{ Raffel_Shazeer_Roberts_etal_}. 

Despite these advancements, model merging approaches for computer vision have been largely confined to simplified settings, such as solely classification tasks, which do not fully reflect the complexities of real-world multi-task applications \citep{Liu_Li_Kuang_etal_2020,liu2021conflict}. Dense prediction tasks such as semantic segmentation and depth estimation, introduce diverse objectives and heterogeneity that challenge the assumptions of existing methodologies \citep{Standley_Zamir_Chen_etal_2020}. Furthermore, existing model merging methods typically assume a constrained setting where each task operates on distinct inputs, whereas real-world applications often follow a single-input, multiple-output (SIMO) paradigm—e.g., predicting both semantic segmentation and depth from the same image—posing new challenges due to shared representations and interdependent task outputs.
\looseness=-1

In this paper, we extend model merging to tackle SIMO multi-task learning settings with a focus on scene understanding tasks. First, we show that existing model merging methodologies, such as Task Arithmetic \citep{ilharco2023task} or TIES \citep{yadav2023ties}, fail in this setting and we show qualitative differences between CLIP classification  \citep{ilharco2023task} and our SIMO paradigm.
Our approach begins with independently fine-tuning a vision foundation model paired with its corresponding lightweight head for each task, under distinct task-specific learning objectives. Subsequently, we merge the learned task-specific encoder weights into a unified shared encoder via task arithmetic, while attaching all lightweight task-specific heads. However, we observe significant performance drops due to the misalignment in the feature representations produced by the shared encoder and those of the task-specific encoders, the latter being aligned with their respective task-specific heads. To mitigate these feature distribution shifts and align the representation of the shared encoder with that of the task-specific heads, we introduce Parameter-Efficient Fine-Tuning (PEFT) strategies \citep{hu2022lora,houlsby2019parameter}, leading to substantial improvement in performance.

Our experiments are conducted on three benchmarks that exemplify diverse MTL settings: NYUv2 \citep{silberman2012indoor}, Cityscapes \citep{Cordts2016Cityscapes}, and a subset of  Taskonomy \citep{taskonomy2018}. These datasets span a range of vision tasks, including semantic segmentation, depth estimation, and surface normal prediction, among others. Using a DINOv2 backbone \citep{oquab2024dinov2learningrobustvisual} and lightweight task-specific heads, we compare our approach to traditional MTL and state-of-the-art model merging baselines. We evaluate performance using task-specific metrics, normalized multi-task performance metrics, and visualization of task relationships, providing a comprehensive analysis of our method's capabilities. Notably, our approach achieves competitive or superior performance to traditional multi-task learning while being more flexible and computationally efficient. 
\looseness=-1

Our contributions can be summarized as follows:
\begin{itemize}
    \item Extending model merging to support single-input, multiple-output (SIMO) multi-task learning, addressing challenges of representation misalignment and task conflict via  lightweight mechanisms to align merged encoders with task-specific heads, preserving task performance despite diverse objectives and outputs.
    \item Providing insights into task relationships through task vectors, offering a novel tool for analyzing task compatibility and representation sensitivity in multi-task learning.
    \item Developing a comprehensive evaluation across diverse benchmarks, demonstrating our method’s efficacy in challenging MTL settings while reducing computational overhead.
\end{itemize}

Overall, our approach offers a scalable and efficient alternative to joint fine-tuning, leveraging the availability of task-specific checkpoints.

\section{Related Work}
\label{sec:related work}

\paragraph{Model Merging}
\label{sec:related work:model merging}

Directly editing models in the weight space has gained significant attention, with early works demonstrating that interpolating the weights of independently trained models often results in low-loss paths, preserving functional performance \citep{garipov2018loss, draxler2018essentially, frankle2020linear}. These findings underpin recent advancements in model merging, such as Task Arithmetic, which showed that arithmetic operations among fine-tuned weights can generate scalable multi-task capabilities \citep{ilharco2023task}, while  theoretical insights into task arithmetic  highlight weight disentanglement as a key factor for successful merging \cite{ortiz2023task}. Further improvements include heuristic-guided merging strategies \citep{davari2024model,jin2023dataless}, addressing parameter interference via resolving redundant updates or sign disagreements \citep{yadav2023ties}, preserving critical weights via the Fisher Information Matrix \citep{matena2022merging,Tam_Bansal_Raffel_2023}, setting the merging coefficients with a linearly increasing schedule \citep{wang2025lines}, and randomly dropping and rescaling the task vectors \citep{yu2024language}. Despite the severe performance drop, the task-specific information is still encoded in the multi-task vector \cite{wang2024localizing}. Ada-merging \citep{yang2023adamerging} and aTLAS \citep{zhang2024knowledge} learn the merging coefficients directly from unlabeled and labeled data, respectively. 
However, the computer vision experiments of these works report solely on multi-task classification benchmark using an open-vocabulary model \citep{radford2021learningtransferablevisualmodels}, while our focus lies on more challenging benchmarks of the multi-task learning literature such as scene understanding.
\looseness=-1

\paragraph{Multi-Task Learning}
\label{sec:related work:mtl}

Learning multiple tasks withing a single model has long been a focus of machine learning research \citep{Ruder_2017,Crawshaw_2020}, evolving significantly with the advent of deep learning. 
MTL innovations have been made by two complementary approaches;  architectural modeling to combine several layers into a single cohesive backbone \citep{Misra_Shrivastava_Gupta_etal_2016,Ma_Zhao_Yi_etal_2018,Ruder_Bingel_Augenstein_etal_2019} and optimization techniques \citep{Cipolla_Gal_Kendall_2018,Chen_Badrinarayanan_Lee_etal_2018,yu2020gradient,Liu_Li_Kuang_etal_2020}, focusing on the descent direction of the joint representation in  a standardized shared-bottom \citep{caruana1997multitask} architecture. The descent direction can be computed either directly on the loss level, e.g., employing techniques such as uncertainty weighting~\citep{Cipolla_Gal_Kendall_2018}, adjusting weights based on each task's loss rate of change~\citep{Liu2018EndToEndML}, or task relationship modeling~\citep{liu2022autolambda}, or by randomly weighting the losses \citep{Lin_Feiyang_Zhang_etal_2022}. Alternatively, the loss can be computed on the gradient level \citep{Chen_Badrinarayanan_Lee_etal_2018,Liu_James_Davison_etal_2022}, such as by resolving task gradient conflicts by aligning shared gradient directions~\citep{NEURIPS2020_16002f7a}, projection-based re-weighting~\citep{yu2020gradient}, and conflict-aware gradient adjustments~\citep{liu2021conflict}, enforcing equal projections across all task gradients in the multi-task  descent direction \citep{Liu_Li_Kuang_etal_2020}, casting the gradient combination as a bargaining game \citep{navon2022multi}. Some works have also explored weight interpolation in learning multiple tasks; to parameterize the Pareto Front \citep{dimitriadis2023pareto,dimitriadis2025pareto,rame2024rewarded} or to improve continual learning of tasks \citep{mirzadeh2020linear}.
In contrast to the end-to-end joint training of the aforementioned approaches, we seek to leverage already trained single-task checkpoints to construct a multi-task model.
\looseness=-1

\section{Background and SIMO Problem Statement}
\label{sec:background}

\paragraph{Background and Notation} 
In our SIMO setting we consider dense prediction tasks, such as semantical segmentation and depth estimation, for which we adopt the shared-bottom model architecture \citep{caruana1997multitask}. For a given task $t$, the task-specific model consists of an encoder with parameters $\bmth_t$ and a task-specific prediction head with parameters \( \pphi_t \). To make a prediction, the encoder produces a representation $\calZ = f_{\textrm{enc}}(\xx; \bmth_t)$, which is then processed by the head to generate the prediction $\hat{\yy}_t = g_t(\calZ; \pphi_t)$. 
To obtain a task-specific model for task $t$, we adopt a two-stage fine-tuning process following \citep{kumar2022finetuning,hazimehtask}. In the first stage, a randomly initialized head $\pphi_t$ is trained while keeping the encoder $\bmth_0$ frozen. In the second stage, both the encoder and the head are fine-tuned jointly.

In the multi-task setting, the model consists of a shared encoder $\bmth_{\text{MTL}}$ across tasks and individual task-specific heads $\{\pphi_t\}_{t=1}^\numtasks$, where the overall parameters are denoted as $(\bmth_{\text{MTL}}, \pphi_1, \dots, \pphi_\numtasks)$ assuming a total of $T$ tasks. During inference, the shared encoder produces a representation shared across each task $\calZ_{\text{shared}} = f_{\textrm{enc}}(\xx; \bmth_{\text{MTL}})$, and the corresponding task-specific head is utilized to generate the prediction for $t$-th task as $\hat{\yy}_t = g_t(\calZ_{\text{shared}}; \pphi_t)$.

\paragraph{Constructing Multi-task Model}  
Our goal is to construct a SIMO model capable of performing multiple tasks based on a single input. A common approach to building such a model is \emph{multi-task learning} (MTL), where the parameters $(\bmth_{\text{MTL}}, \pphi_1, \dots, \pphi_\numtasks)$ are trained jointly on the training set of all tasks. However, MTL is computationally expensive, and requires access to all training data.

Recently, \emph{model merging} has emerged as an efficient and flexible alternative to construct a multi-task model. With access to task-specific checkpoints $\{(\bmth_t, \pphi_t)\}_{t=1}^\numtasks$, model merging combines these checkpoints to form a multi-task model with minimal training. Unlike MTL, model merging does not require access to the training data of each task; instead, it directly merges the available individual checkpoints, significantly reducing computational overhead.

Among the model merging methods, \emph{Task Arithmetic} (TA) \citep{ilharco2023task} isolates the effect of fine-tuning on each task by operating in the space of residuals or \textit{task vectors}, defined as 
\(
\tv{t} = \ttheta_t - \ptm, \ \forall t \in [\numtasks],
\)
These task vectors are aggregated into a multi-task vector which captures the information across tasks,
\(
\tvopt = g(\tv{1}, \tv{2}, ..., \tv{T}),
\)
where $g$ is an aggregation function. In task arithmetic, $g$ is a simple summation. The final multi-task encoder weights are then obtained as  
\(
\ttheta_{\text{MTL}} = \ptm + \alpha \tvopt,
\)
where \( \alpha > 0 \) is a scaling factor tuned on a held-out validation set.
More recently, \emph{Ties-Merging} (TIES) \citep{yadav2023ties} follows a similar procedure to task arithmetic, while improving the merging function $g$ to reduce the parameter interferences during merging. 
However, these approaches have largely been confined to simplified classification tasks with disjoint label spaces, neglecting more challenging settings like dense prediction. Such tasks, including semantic segmentation and depth estimation, involve richer supervision and spatially structured outputs that introduce greater heterogeneity and inter-task dependencies.
\looseness=-1

\section{Model Merging Fails in SIMO Setting}

\begin{figure}[t]
    \centering
    \includegraphics[width=0.8\linewidth]{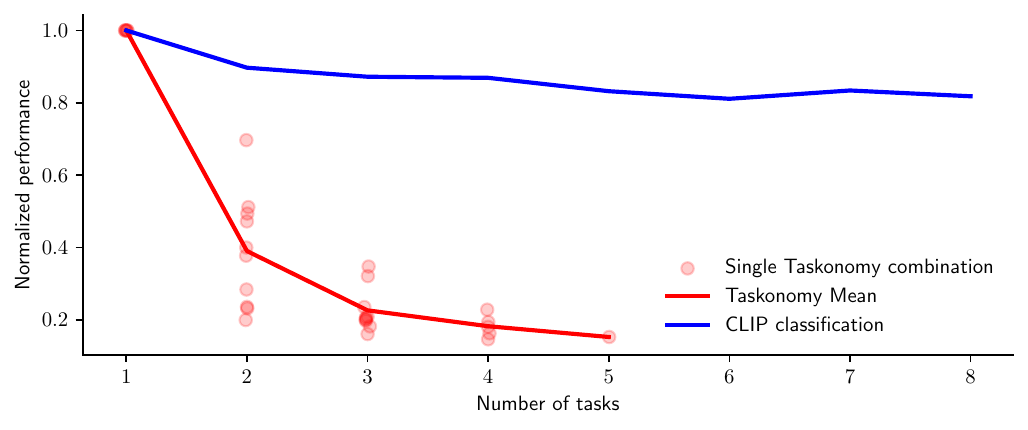}
    \caption{Comparison of performance deterioration between dense prediction (Taskonomy \citep{taskonomy2018} with 5 tasks) and vision classification (8-task benchmark introduced in \citep{ilharco2023task}) benchmarks as a function of the number. Each point corresponds to normalized performance of the Task Arithmetic \citep{ilharco2023task} baseline for a $k$-task combination. Dense prediction combinations exhibit a steeper decrease compared to vision classification, indicating the increased difficulty of the setting.}
    \label{fig:comparison}
\end{figure}

In this section, we evaluate the effectiveness of applying model merging methods directly to construct SIMO models and show the significant performance drop of existing methods in this setting. 
We consider merging task-specific checkpoints on the Taskonomy dataset \citep{taskonomy2018}, a multi-task benchmark consisting of 5 dense-prediction tasks.
Each task-specific checkpoint is fine-tuned from a DINOv2 base model \citep{oquab2024dinov2learningrobustvisual} following the 2-stage fine-tuning process described in \autoref{sec:background}.

After fine-tuning, we obtain the single-task checkpoints $\{(\bmth_t,\pphi_t)\}_{t=1}^\numtasks$, from which we construct a SIMO model with parameters $(\bmth_{\text{MTL}}, \pphi_1,\dots,\pphi_\numtasks)$ using task arithmetic. 
We evaluate the normalized performance of the merged model, i.e., the percentage of performance retention w.r.t. the single-task checkpoints, while varying number of tasks being merged in Taskonomy. 
\autoref{fig:comparison} presents the normalized performance\footnote{Details on the normalized performance in Appendix~\ref{app:metrics}.} of the merged model against the number of task-specific checkpoints being merged, where we observe a drastic performance drop as number of tasks increases. Notably, with only 2 task-specific models being merged, the performance of task arithmetic drops by around 60\% compared to the individual models, while merging the checkpoints from a total of 5 tasks results in less than 20\% normalized performance. The failure of task arithmetic in this SIMO setting is in stark contrast to its success in other settings, such as merging CLIP-based image classification models \citep{radford2021learningtransferablevisualmodels}, where it preserves almost full performance when merging 2 task-specific checkpoints \citep{ilharco2023task}.

\section{SIMO Model Merging with Feature Re-alignment}
\label{sec:realignment}

To understand the collapsing performance of model merging methods, we start with comparing the difference in the settings primarily considered in previous model merging works and the SIMO setting.
Model merging methods, such as Task Arithmetic \citep{ilharco2023task} and Ties-merging \citep{yadav2023ties}, have primarily been applied in the context of open-vocabulary models like CLIP models \citep{radford2021learningtransferablevisualmodels}. CLIP's architecture consists of a visual encoder and a text encoder that jointly learn an aligned embedding space. During fine-tuning, only the visual encoder is adapted to the target task, while the text encoder is frozen and serves as a classification head for making predictions.
As a result, all individual encoders and the merged encoder remain in a consistent embedding space which is aligned with the pre-trained text encoder serving as a universal decoder for making predictions on each task. It allows the merged visual encoder to generalize across tasks without the need for explicit representation alignment for each task.

However, this assumption breaks in our SIMO setting, where each task requires an independent task-specific head that is fine-tuned alongside the encoder. In the SIMO setting, merging individual task-specific encoders produces an encoder that does not lie in the same embedding space as the ones originally used to train the task-specific decoders. This representation mismatch or \textit{bias} \cite{representation_surgery} leads to degraded performance, as the merged encoder produces representations misaligned with the expected input distributions of the task-specific heads. Consequently, naive model merging methods fail to generalize effectively across multiple tasks in SIMO scenarios.

To alleviate the issue of representation mismatch in model merging for SIMO settings, we propose two strategies, re-aligning the head and re-aligning the joint representation.
It is important to note that the single-task checkpoints have been fine-tuned on task-specific train datasets $\calD^t_{\textrm{train}}, \forall t\in[\numtasks]$, while our strategies operate by assuming access to a much smaller \textit{multi-task} validation dataset $\calD_{\textrm{val}}^{1:\numtasks}$.
\looseness=-1

\subsubsection{Head Re-alignment}
The merged encoder is frozen and the task-specific heads are fine-tuned separately on each task to adapt to the new representation, which allows for aligning in parallel and without the challenges of joint learning. This approach is computationally efficient; given sample $(\xx, \yy)\in\calD_{\textrm{val}}^{t}$ of dataset $t$, the encoder representation $\calZ=f_{\textrm{enc}}(\xx; \bmth)$ can be pre-computed once since fine-tuning occurs solely on the task-specific heads.

\subsubsection{Representation Re-alignment} 
\begin{figure}[t]
    \centering
    \includegraphics[width=\linewidth]{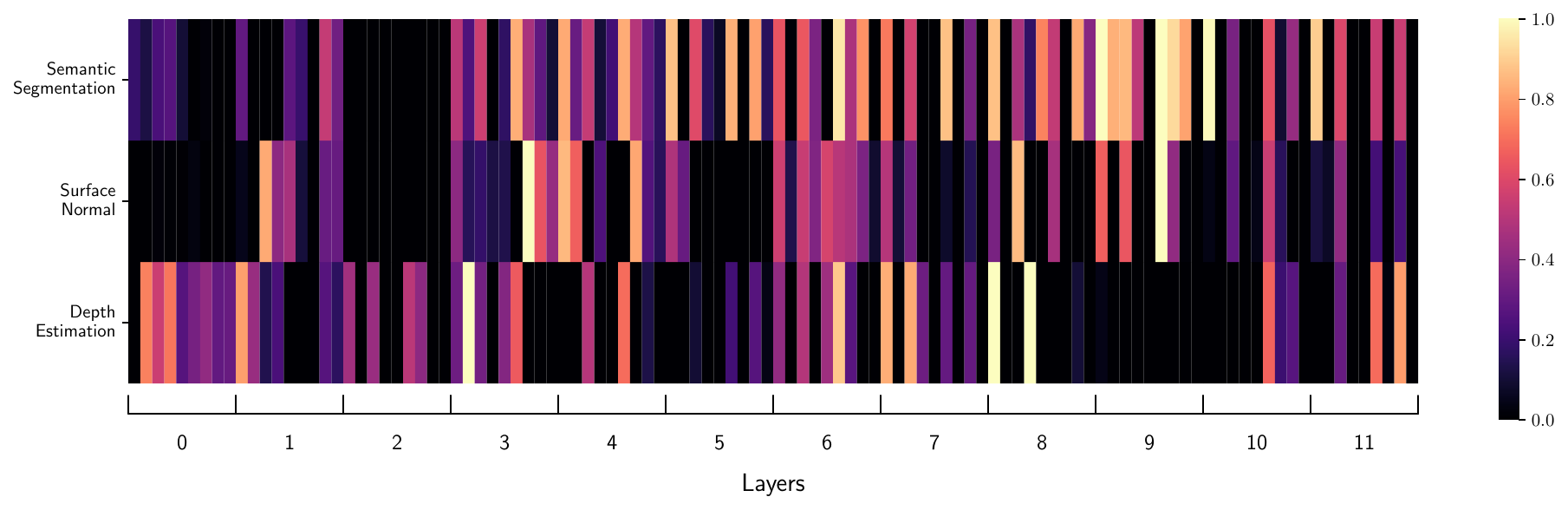}
    \caption{Optimal merging coefficients per layer as found by Adamerging \cite{yang2023adamerging} on NYUv2 for the DINOv2 architecture consisting of 12 repeating blocks of layers. Tasks deem important different parts of the representation; therefore applying computationally tractable but simple merging approaches like Task Arithmetic results in a representation misalignment in the encoder. }
    \label{fig:adamerging}
\end{figure}

Fine-tuning solely the head resolves the issue of the misalignment in the representation expected by the task-specific heads caused by merging. However, the encoder representation is still distorted, resulting in lower quality embeddings that can affect the final performance, especially when the number of tasks and hence the representation bias grows. 
For analysis purposes, we consider anisotropic scaling for aggregating task vectors \cite{yang2023adamerging,zhang2024knowledge}, where a different scaling coefficient per task and per layer is obtained via backpropagation.
Differently than CLIP classification where a consistent increasing trend over the layer number is observed across tasks \cite{wang2025lines}, \autoref{fig:adamerging} shows that dense prediction tasks utilize the encoder representation differently. Therefore, merging with a global coefficient as in Task arithmetic, distorts the representation and requires a more targeted approach.

After merging the encoders from different tasks, we employ a Parameter-Efficient Fine-Tuning \citep{houlsby2019parameter} approach to align the encoded weights of each layer towards a joint representation. Specifically, we consider a light-weight LoRA module \cite{hu2022lora} after each repeated block in the transformer architecture \cite{vaswani2017attention}.  Assume that each block implements a function $f: \bbR^n\mapsto\bbR^m$. We then augment the architecture with a low-rank adapter per block, i.e., trainable matrices $\AA\in\bbR^{r\times n}$ and $\BB\in\bbR^{m\times r}$ for $r\lll\min\{n, m\}$. For input $\xx\in\bbR^n$, the modified output of the block is $f_{\textrm{modified}}(\xx)=f(\xx) + \BB\AA\xx$. 
\looseness=-1

By introducing these alignment techniques, we enable model merging to more effectively function in multi-task SIMO settings, overcoming the representation shift that arises due to task-specific decoder training. In the following section, we demonstrate the effectiveness of the re-alginment on several benchmarks and analyze their effectiveness in improving the performance of traditional model merging methods.
\looseness=-1

\begin{table}[t]
    \centering
    \caption{Performance on NYUv2 using the DINOv2-base + linear head architecture, reported as absolute values with normalized metrics in parentheses and overall relative multi-task performance ($\Delta_{MTL}$).}
    \label{tab:nyuv2}
    \begin{tabular}{@{}ccccc@{}}
    \toprule
    \multirow{2}{*}{\textbf{Method}} & \textbf{Sem. Seg.} & \textbf{Depth} & \textbf{Surface Normal} & \multirow{2}{*}{\textbf{$\Delta_{MTL} \uparrow$}} \\ 
    & [mIoU $\uparrow$] & [aErr. $\downarrow$] & [mDist. $\downarrow$] & \\ 
    \midrule
    Task-Specific & 70.79 & 0.2314 & 18.02 & - \\ 
    MTL & $69.01_{(97.5\%)}$ & $0.2531_{(91.4\%)}$ & $19.01_{(94.8\%)}$ & -5.4\% \\ 
    Task Arithmetic & $65.78_{(92.9\%)}$ & $0.4171_{(55.5\%)}$ & $26.13_{(69.0\%)}$ & -27.5\% \\ 
    TIES & $62.62_{(88.5\%)}$ & $0.4157_{(55.7\%)}$ & $31.38_{(57.4\%)}$ & -32.8\% \\ 
    \midrule
    TA + head re-align (Ours) & $69.95_{(98.8\%)}$ & $0.2524_{(91.7\%)}$ & $19.29_{(93.5\%)}$ & -5.4\% \\
    TA + repr. re-align (Ours) & $68.95_{(97.4\%)}$ & $0.2539_{(91.1\%)}$ & $18.99_{(94.9\%)}$ & -5.5\% \\
    TIES + head re-align (Ours) & $66.25_{(93.6\%)}$ & $0.2764_{(83.7\%)}$ & $18.65_{(96.6\%)}$ & -8.7\% \\
    TIES + repr. re-align (Ours) & $68.49_{(96.8\%)}$ & $0.2576_{(89.8\%)}$ & $18.99_{(94.9\%)}$ & -6.2\% \\
    \bottomrule
    \end{tabular}
\end{table}

\section{Experiments}
\label{sec:experiments}

In this Section, we empirically evaluate the effectiveness of our proposed approaches to adapt model merging methods to the SIMO setting. We consider two strong baselines in model merging literature, including task arithmetic \citep{ilharco2023task} and Ties-merging \citep{yadav2023ties}, highlighting the performance improvement over them after applying our feature re-alignment methods. For the evaluation metrics, we apply different metrics for each reported task, and also report the normalized performance of the multi-task model normalized by task-specific model's performance. As an overall assessment of the multi-task performance, we report also overall relative multi-task performance \(\Delta_{MTL}\), measuring the relative performance drop of the multi-task model. Appendix~\ref{app:metrics} details the evaluation metrics.
\looseness=-1

\subsection{Experimental Setup}
\label{sec:experimental setup}

\paragraph{Benchmarks} We consider three multi-task benchmarks:
\begin{itemize}
    \item NYUv2 \citep{silberman2012indoor}: 13-class semantic segmentation, monocular depth estimation, and surface normal prediction;
    \item Cityscapes \citep{Cordts2016Cityscapes}: 19-class semantic segmentation, 10-class part segmentation~\citep{de2021part}, and disparity (inverse depth) estimation
    \item Taskonomy \citep{taskonomy2018}: Autoencoder (image compression and decompression), monocular depth estimation, surface normal prediction, SURF keypoints detection, and canny edge detection. For computational reasons, we consider a small subset of the original Taskonomy dataset, similar to \cite{bachmann2022multimae}. 
\end{itemize}

The details of the evaluated benchmarks as well as the evaluation metrics are provided in Appendix~\ref{app:benchmarks}.

\paragraph{Model Architecture} We adopt DINOv2-base (86.6M parameters) as our VFM. For the multi-head module, we employ lightweight architectures, specifically either a linear head ($\approx 0.1$M parameters) for NYUv2, or a DPT decoder ($\approx3.5$M parameters) \citep{ranftl2021visiontransformersdenseprediction} for Cityscapes and Taskonomy. Our setup follows \cite{oquab2024dinov2learningrobustvisual} for dense recognition tasks.

\paragraph{Training Details} The single-task models required for our proposed approach, as well as the MTL models used as baselines, were trained in a two-phase process following a “probing then full fine-tuning” paradigm~\citep{kumar2022finetuning}. In the first phase, the task-specific heads were trained until convergence while keeping the backbone frozen. The second phase conducted joint fine-tuning with the backbone unfrozen until validation loss plateaued. Single-task models employed the cross-entropy loss for semantic segmentation, the dot product loss for surface normal, and the $L_1$ loss for all other tasks. For multi-task training, we use an unweighted mean of the losses for the individual tasks \cite{caruana1997multitask}. Further details regarding the hyperparameter search are provided in Appendix~\ref{app:training_details}.
\looseness=-1

\paragraph{Data Access} Let $\calD=\calD^{1:\numtasks}=\left\{\left(\xx^{(i)}, \yy_1^{(i)}, \dots, \yy_\numtasks^{(i)}\right)\right\}_{i=1}^N$ be a multi-task dataset for $\numtasks$ tasks. The multi-task baseline and single-task checkpoints have been obtained by fine-tuning on $\calD_{\textrm{train}}$; single-task training has access to only one of the targets per task. Our approaches assume access to these single-task models and to a smaller multi-task validation dataset $\calD_{\textrm{val}}$, with  $|\calD_{\textrm{val}}| \lll |\calD_{\textrm{train}}|$.

\subsection{Experimental Results}
We empirically demonstrate the effectiveness of our proposed solutions for enhancing model merging techniques in SIMO settings. For each benchmark, we report the performance of the model merging methods before and after our proposed solutions, and also report the performance of single-task models and multi-task learning model as baselines. 

\paragraph{Results on NYUv2} \autoref{tab:nyuv2} presents the results on NYUv2.  Task Arithmetic and TIES exhibit substantial performance degradation when merging task-specific models with a relative multi-task performance drop of \(-27.5\%\) and \(-32.8\%\), respectively. The degradation primarily stems from misalignment between the merged encoder and the original task-specific heads, which were optimized for individual fine-tuned models. As we see from \autoref{tab:nyuv2}, simply applying head realignment (TA + head re-alignment, TIES + head re-alignment) significantly mitigates this issue, boosting \(\Delta_{MTL}\) to normalized performance loss of only \(-5.4\%\) and \(-8.7\%\), respectively. 
With our proposed feature re-alignment strategies to existing model merging methods, their performance is boosted to the same level as MTL (with \(\Delta_{MTL}\) of \(-5.4\%\))  with lower computational costs and without the need to train on a specific task combination chosen a prior.
\looseness=-1

\begin{table}[t]
    \centering
    \vspace{0.5cm}
    \caption{Performance on Cityscapes using the DINOv2-base + DPT head architecture, reported as absolute values with normalized metrics in parentheses and overall relative multi-task performance ($\Delta_{MTL}$).}
    \label{tab:cityscapes}
    \begin{tabular}{@{}ccccc@{}}
    \toprule
    \multirow{2}{*}{\textbf{Method}} & \textbf{Sem. Seg.} & \textbf{Part Seg.} & \textbf{Disparity} & \multirow{2}{*}{\textbf{$\Delta_{MTL} \uparrow$}} \\ 
    & [mIoU $\uparrow$] & [mIoU. $\uparrow$] & [aErr. $\downarrow$] & \\ 
    \midrule
    Task-Specific & 58.76 & 50.88 & 0.0102 & - \\ 
    MTL & $56.36_{(95.9\%)}$ & $50.36_{(99.0\%)}$ & $0.0108_{(94.4\%)}$ & -3.7\% \\ 
    Task Arithmetic & $33.56_{(57.1\%)}$ & $40.55_{(79.7\%)}$ & $0.0147_{(69.4\%)}$ & -31.3\% \\ 
    TIES & $30.45_{(51.8\%)}$ & $39.91_{(78.4\%)}$ & $0.0111_{(91.8\%)}$ & -26.1\% \\ 
    \midrule
    TA + head re-align (Ours) & $53.36_{(90.8\%)}$ & $48.94_{(96.2\%)}$ & $0.0105_{(97.2\%)}$ & -5.3\% \\
    TA + repr. re-align (Ours) & $54.59_{(92.9\%)}$ & $49.37_{(97.0\%)}$ & $0.0105_{(96.7\%)}$ & -4.5\% \\
    TIES + head re-align (Ours) & $52.13_{(88.7\%)}$ & $48.32_{(95.0\%)}$ & $0.0103_{(99.4\%)}$ & -5.7\% \\
    TIES + repr. re-align (Ours) & $53.12_{(90.4\%)}$ & $48.42_{(95.2\%)}$ & $0.0104_{(98.1\%)}$ & -5.5\% \\
    \bottomrule
    \end{tabular}
\end{table}

\paragraph{Results on Cityscapes}
We observe a similar trend on Cityscapes in \autoref{tab:cityscapes}. Direct merging via TA and TIES results in substantial performance losses with \(\Delta_{MTL}\) of \(-31.3\%\) and \(-26.1\%\), while re-aligning task-specific heads post-merging leads to notable improvements, with TA + head re-alignment and TIES + head re-alignment improving performance loss to only \(-5.3\%\) and \(-5.7\%\), respectively. 
The introduction of lightweight adapters to re-align the overall representation further improves the performance of task arithmetic to  \(-4.5\%\), very close to that achieved by MTL (\(-3.7\%\)).

\paragraph{Results on Taskonomy}
We finally evaluate the performance of our proposed methods on Taskonomy, a more challenging setting due to its diverse range of tasks. The number of tasks has increased from 3 to 5, compounding on the feature misalignment after merging as seen in \autoref{fig:comparison}.
As we observe from \autoref{tab:taskonomy}, Task arithmetic exhibits extreme performance degradation (\(\Delta_{MTL} = -83.3\%\)), while TIES also struggles at \(-78.0\%\). 
While the vanilla model merging methods lead to collapsing performance, our proposed feature re-alignment methods significantly mitigate the performance drop.
Head re-alignment significantly improves the performance of task arithmetic and TIES to a \(\Delta_{MTL}\) of \(-36.0\%\) and \(-31.1\%\), respectively. Furthermore, representation re-alignment yields larger improvements, with TIES + representation re-alignment achieving the best performance at \(-17.4\%\), notably surpassing the performance of MTL with \(-25.7\%\) despite much less computational costs. These findings suggest that while task diversity amplifies feature misalignment after merging, our proposed modifications such as post-hoc head fine-tuning and lightweight feature adaptation can substantially mitigate the issue. We refer to Appendix~\ref{app:qualitative_results} for some qualitative results on the Taskonomy dataset.
\looseness=-1

\begin{table}[t]
    \centering
    \vspace{0.5cm}
    \caption{Performance on Taskonomy using the DINOv2-base + DPT head architecture, reported as absolute values with normalized metrics in parentheses and overall relative multi-task performance ($\Delta_{MTL}$).}
    \label{tab:taskonomy}
    \resizebox{\linewidth}{!}{%
    \begin{tabular}{@{}ccccccc@{}}
    \toprule
    \multirow{2}{*}{\textbf{Method}} & \textbf{Autoencoder} & \textbf{Depth} & \textbf{Surface Normals} & \textbf{Edge Texture} & \textbf{Keypoints} & \multirow{2}{*}{\textbf{$\Delta_{MTL} \uparrow$}} \\ 
    & [aErr. $\downarrow$] & [aErr. $\downarrow$] & [aErr. $\downarrow$] & [aErr. $\downarrow$] & [aErr. $\downarrow$] & \\ 
    \midrule
    Task-Specific & 0.0250 & 0.0146 & 0.0696 & 0.0126 & 0.0062 & - \\ 
    MTL & $0.0324_{(77.2\%)}$ & $0.0170_{(85.9\%)}$ & $0.0772_{(90.2\%)}$ & $0.0213_{(59.2\%)}$ & $0.0105_{(59.0\%)}$ & -25.7\% \\ 
    Task Arithmetic & $0.2067_{(12.1\%)}$ & $0.0668_{(21.9\%)}$ & $0.4467_{(25.6\%)}$ & $0.0750_{(16.8\%)}$ & $0.0846_{(7.3\%)}$ & -83.3\% \\ 
    TIES & $0.3132_{(8.0\%)}$ & $0.3000_{(4.9\%)}$ & $0.0936_{(74.4\%)}$ & $0.1211_{(10.4\%)}$ & $0.0513_{(12.1\%)}$ & -78.0\% \\ 
    TALL-Mask & $0.0650_{(38.5\%)}$ & $0.2271_{(6.4\%)}$ & $0.0768_{(90.6\%)}$ & $0.0782_{(16.1\%)}$ & $0.0196_{(31.6\%)}$ & -63.4\% \\ 
    \midrule
    TA + head re-align (Ours) & $0.0301_{(83.1\%)}$ & $0.0270_{(54.1\%)}$ & $0.1428_{(48.7\%)}$ & $0.0256_{(49.2\%)}$ & $0.0073_{(84.9\%)}$ & -36.0\% \\
    TA + repr. re-align (Ours) & $0.0283_{(88.3\%)}$ & $0.0186_{(78.5\%)}$ & $0.0968_{(71.9\%)}$ & $0.0199_{(68.3\%)}$ & $0.0071_{(87.3\%)}$ & -21.1\% \\
    TIES + head re-align (Ours) & $0.0266_{(94.0\%)}$ & $0.0290_{(50.3\%)}$ & $0.1707_{(40.8\%)}$ & $0.0213_{(59.2\%)}$ & $0.0060_{(100.3\%)}$ & -31.1\% \\
    TIES + repr. re-align (Ours) & $0.0283_{(88.3\%)}$ & $0.0167_{(87.4\%)}$ & $0.0903_{(77.1\%)}$ & $0.0195_{(64.6\%)}$ & $0.0065_{(95.4\%)}$ & -17.4\% \\
    \bottomrule
    \end{tabular}
    }
\end{table}

\subsection{Reduced Computational Efficiency}

Multi-task learning involves jointly optimizing over diverse losses, introducing trade-offs \citep{Standley_Zamir_Chen_etal_2020,Fifty_Amid_Zhao_etal_2021,liu2022autolambda} and requiring an a priori selection of the task combination. Our approach leverages the existence of single task checkpoints, which are more straightforward to optimize, and employs a post-training alignment procedure on the \textit{multi-task} validation dataset $\calD_{\textrm{val}}^{1:\numtasks}$ that is computationally more tractable than multi-task training from scratch on the \textit{multi-task} train dataset $\calD_{\textrm{train}}^{1:\numtasks}$.

The head realignment method leverages a frozen encoder, enabling to first pre-compute the feature representations across the dataset. Subsequent training only involves forward passes through the task-specific heads, substantially reducing computational overhead. 

Although our proposed Representation Re-alignment technique requires joint training of the SIMO model using a MTL objective, only the lightweight modules within the encoder and task-specific heads are trained. Incorporating a PEFT module introduces only a negligible increase in the parameter count, as shown in \autoref{tab:efficiency},  and reduces the computational demands while training. Nevertheless, at inference time, merging the adapter weights into the encoder effectively eliminates this overhead, optimizing performance.

\begin{table}[t]
    \centering
    \caption{Number of trainable parameters (in millions) required for each technique.}
    \label{tab:efficiency}
    \begin{tabular}{l|c|c|c}
    \toprule
    \multirow{2}{*}{\textbf{Method}} & \textbf{NYUv2} & \textbf{Cityscapes} & \textbf{Taskonomy} \\ 
    & Params. & Params. & Params.\\  
    \midrule
    MTL                 & 88.3 M & 95.9 M & 103.9 M \\  
    TA + head re-align   & 1.7 M & 9.3 M & 17.3 M \\ 
    TA + repr. re-align & 2.0 M & 9.6 M & 17.6 M \\
    \bottomrule
    \end{tabular}
\end{table}

\subsection{Identifying Task Relationships} \label{sec:task_relation}
As a byproduct of leveraging task vectors to construct our SIMO models, we propose a simple yet novel method for identifying task relationships, e.g., for observing which tasks can be mutually beneficial or interfering. Given a primary task $t$, we aim to understand how adding its task-specific information to a pre-trained model affects other tasks. To achieve this, we initialize a SIMO model comprising of a pre-trained vision encoder $\bmth_{0}$ and fine-tuned task-specific heads $(\pphi_1, \dots, \pphi_\numtasks)$, assuming a total of $T$ tasks. We then perturb the shared encoder by adding information from a single primary task $\lambda \cdot \tau_{t}$, where $\lambda$ is chosen (e.g., $\lambda \in [0, 1.5]$). The core idea is to then evaluate how task-specific metrics vary with $\lambda$, perturbing the encoder’s feature representation toward task $t$, thereby exposing alignment or conflicts among the different $T-1$ tasks.

\begin{figure}[t]
    \centering
    \includegraphics[width=\linewidth]{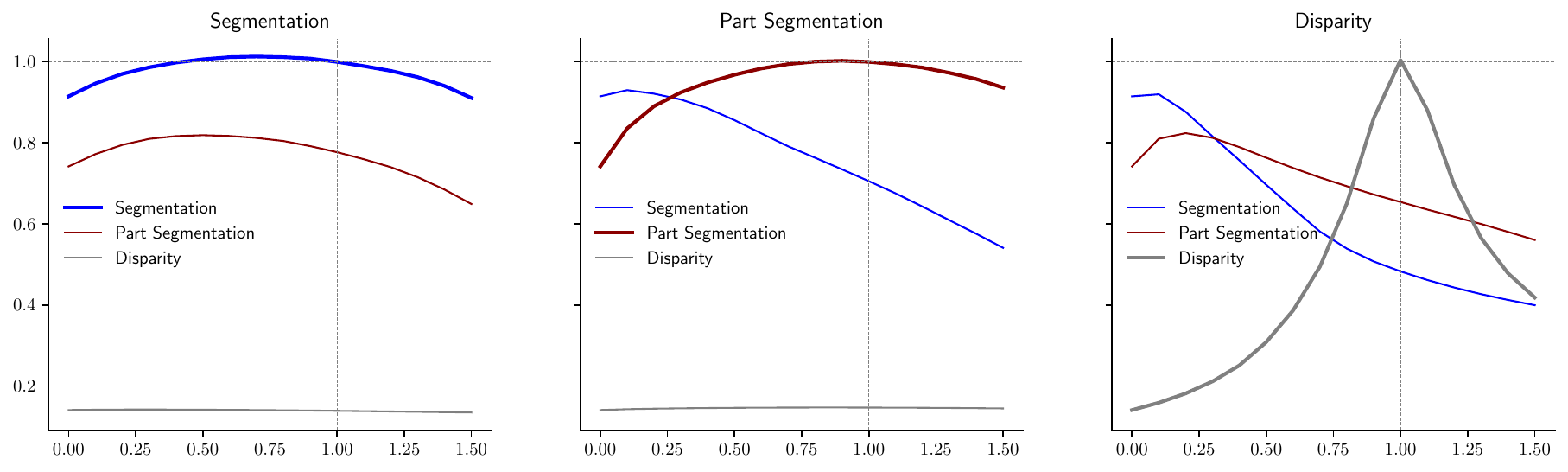}
    \caption{Visualization of task relationships for Cityscapes. Each plot presents the normalized performance on each dataset as a function of the task vector added to the model with a coefficient of the x-axis, forming the model $\ptm+\lambda \bm{\tau}$, for $\bm{\tau}\in\{\tv{seg}, \tv{part\_seg},\tv{disp}\}$.}
    \label{fig:cityscapes_task_relationships}
\end{figure}

Figure~\ref{fig:cityscapes_task_relationships} illustrates our approach by analyzing task relationships within the Cityscapes dataset, providing two key insights:
\begin{enumerate}
    \item \textit{task relationships}: we assess how the performance of secondary tasks evolves as the primary task is introduced. Beneficial interactions emerge when the performance of the secondary task improves, e.g., semantic segmentation enhances part segmentation. In contrast,  conflicting interactions results in consistent degradation of secondary task performance, e.g., part segmentation undermining semantic segmentation; 
    \item \textit{sensitivity of the feature representations}: certain tasks, such as disparity, show high sensitivity to perturbations in the encoder weights, requiring $\lambda \approx 1$ to recover optimal performance. Therefore, naively merging such sensitive weights may result in information loss. 
\end{enumerate}
These observations provide critical insights into task compatibility and their potential for effective multi-task learning. Further visualizations of task relationships for the NYUv2 and Taskonomy dataset are provided in Appendix~\ref{app:task_relationships}.

\section{Conclusion}
\label{sec:conclusion}
This study demonstrates that single-input multiple-output models can be efficiently built by leveraging single-task checkpoints and properly applying model merging techniques. By mitigating representation bias in the encoder and aligning task-specific heads to the new feature representation, our approach achieves competitive performance at a lower computational cost compared to standard multi-task fine-tuning. Experiments on large scale benchmarks, such as NYUv2, Cityscapes and Taskonomy confirm the effectiveness of the proposed method.
These results suggest a promising alternative to expensive joint training, enabling scalable and flexible multi-task systems that can be assembled post hoc from existing models.
\looseness=-1

\bibliography{refs_with_url}

\clearpage
\appendix

\section{Details on evaluation benchmarks}
\label{app:benchmarks}
\begin{itemize}
    \item NYUv2: the training set comprises 795 samples with an image size of $280 \times 378$, while the validation set contains 400 samples and the test set includes 254 samples, both maintaining the same image dimensions.
    \item Cityscapes: the training set consists of 2,380 samples with an image size of $128 \times 256$, the validation set includes 595 samples, and the test set comprises 500 samples, all with identical dimensions.
    \item Our small subset of the Taskonomy dataset: the training set contains 16,000 samples, drawn from eight buildings (allensville, coffeen, collierville, leonardo, merom, pinesdale, ranchester, stockman) with a maximum of 2,000 samples per building and an image size of $256 \times 256$. The validation and test sets each include 4,000 samples, sourced from four buildings each (beechwood, corozal, klickitat, shelbyville for validation; ihlen, mcdade, onaga, tolstoy for test), with a maximum of 1,000 samples per building, maintaining the same image dimensions. 
\end{itemize}

\section{Evaluation Metrics} 
\label{app:metrics}

\paragraph{Task-specific metrics}
We assess the performance of our dense prediction tasks across all datasets using three primary metrics: (1) mean Intersection over Union (mIoU), which quantifies the overlap between predicted and ground-truth regions, (2) mean angular distance (mDist), which measures the average angular deviation (in degrees) between predicted and ground-truth values, and (3) absolute error (aErr), defined as the L1 difference between predictions and ground truth. For NYUv2 and Cityscapes we evaluate segmentation tasks via mIoU, surface normals via mDist., and depth/disparity estimation via aErr. For the Taskonomy dataset, all tasks are evaluated using the aErr. 

\paragraph{Normalized performance}
Assuming limited data availability and computational budget,  and similar to prior model merging works \citep{yadav2023ties,yu2024language}, we measure the drop in performance relative to the single-task case as a proxy for retention. Formally, let $M$ denote a given metric for task $t$ from the model $m$ being evaluated, and $b$ represents the single-task baselines.
\begin{equation}
    \tilde{M}=\frac{1}{T} \sum_{t=1}^T \left( \frac{M_{m, t}}{M_{b, t}} \right)^{(-1)^{l_t}} \label{eq:normalize} 
\end{equation}
where $l_t$ is set to $1$ if lower values of $M_{m,t}$ indicate better performance, and 0 otherwise. 

\paragraph{Relative multi-task performance}

Following \cite{wang2024localizing,maninis2019attentivesingletaskingmultipletasks}, we also report the overall relative multi-task performance $\Delta_{MTL}$, which measures the relative performance drop of the multi-task model compared to the individual single-task models:
\begin{equation}
    \Delta_{\mathrm{MTL}} = \frac{1}{T} \sum_{t=1}^T (-1)^{l_t} \frac{M_{m, t}-M_{b, t}}{M_{b, t}} \label{eq:delta_mtl}
\end{equation}
Overall, our objective is to minimize the metrics $\tilde{M}$ and $ \Delta_{\mathrm{MTL}}$.

\section{Training Details} \label{app:training_details}
We detail the training hyperparameters used across all experiments, employing DINOv2-B as the encoder with the following output layer $\{3, 6, 9, 12\}$ + CLS token used by the task-specific heads (linear projections or DPT architecture) for the NYUv2, Cityscapes, and Taskonomy datasets. The adopt the AdamW optimizer with a batch size of 16 for NYUv2 and Cityscapes, and 32 for Taskonomy. The learning rate hyperparameter tuning is conducted via a Bayesian optimization technique with a log-uniform distribution ranging from $1 \times 10^{-8}$ to $1 \times 10^{-3}$. We employ a once-cycle learning rate scheduler, with a warm-up ratio of 0.05 and gradient clipping set to 25.0. Our training includes a maximum of 50 epochs for task-specific models, 200 epochs for multi-task learning, and 200 epochs for our proposed representation re-alignment approaches.

\section{Qualitative Results} \label{app:qualitative_results}
\begin{figure}[t]
    \centering
    \includegraphics[width=\linewidth]{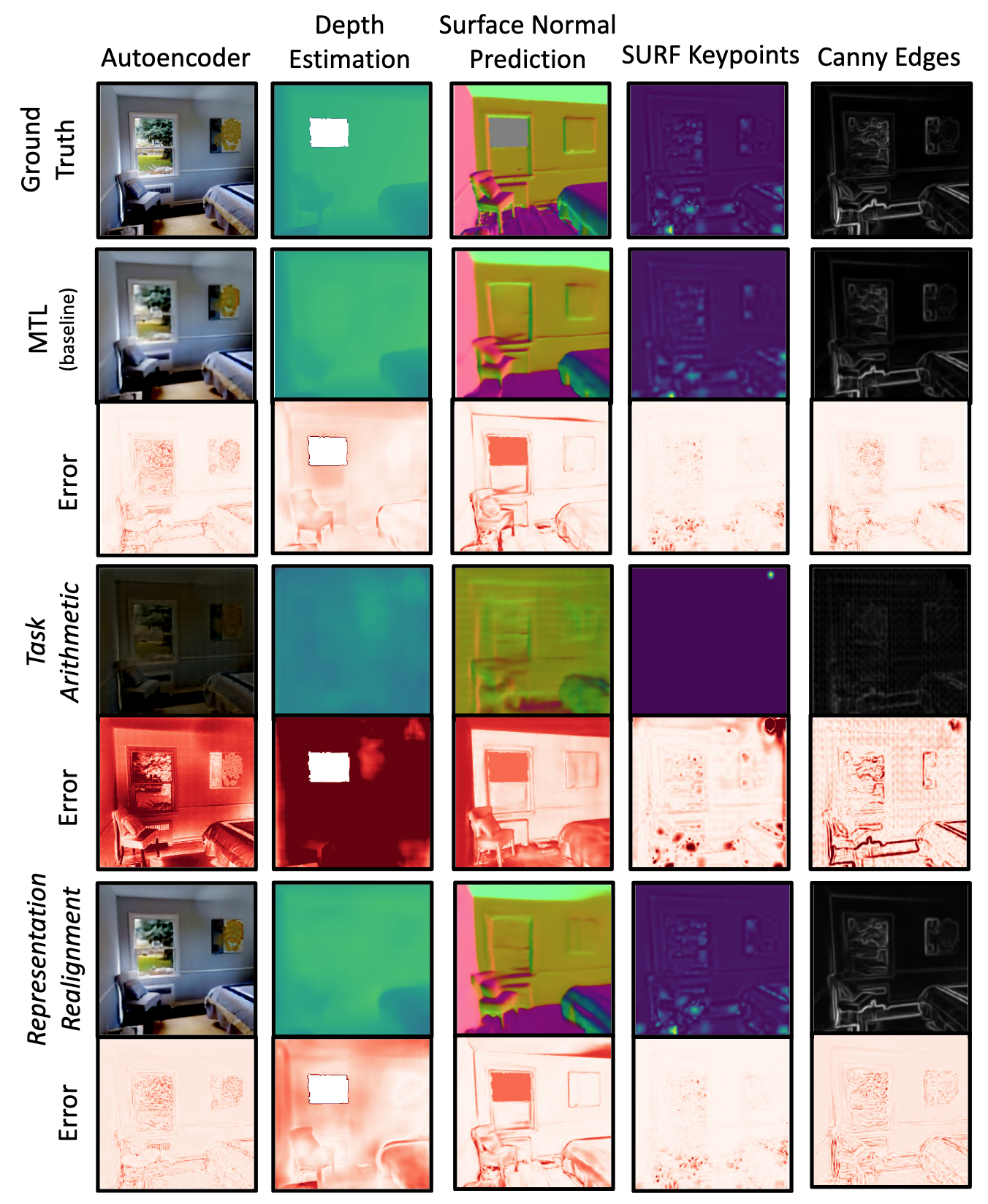}
    \caption{Qualitative results comparing MTL, Task Arithmetic, and our Representation Re-alignment technique on the Taskonomy dataset.}
    \label{fig:taskonomy_qualitatively}
\end{figure}

Figure~\ref{fig:taskonomy_qualitatively} visualizes the results of MTL (serving as our baseline), standard model merging techniques (e.g. Task Arithmetic), and our proposed representation re-alignment approach, with our figure clearly illustrating the shortcomings of standard model merging techniques. Although Table~\ref{tab:taskonomy} quantitatively suggests that MTL and our approach may appear suboptimal based on normalized and relative multi-task performance metrics, these visualizations highlight that one should not consider them as ground truths. Highlighting the limitations of using the absolute error (aErr) commonly used for the Taskonomy dataset to report multi-task dynamics.

\section{Visualizing Task Relationships} \label{app:task_relationships}
\begin{figure}[t]
    \centering
    \includegraphics[width=\linewidth]{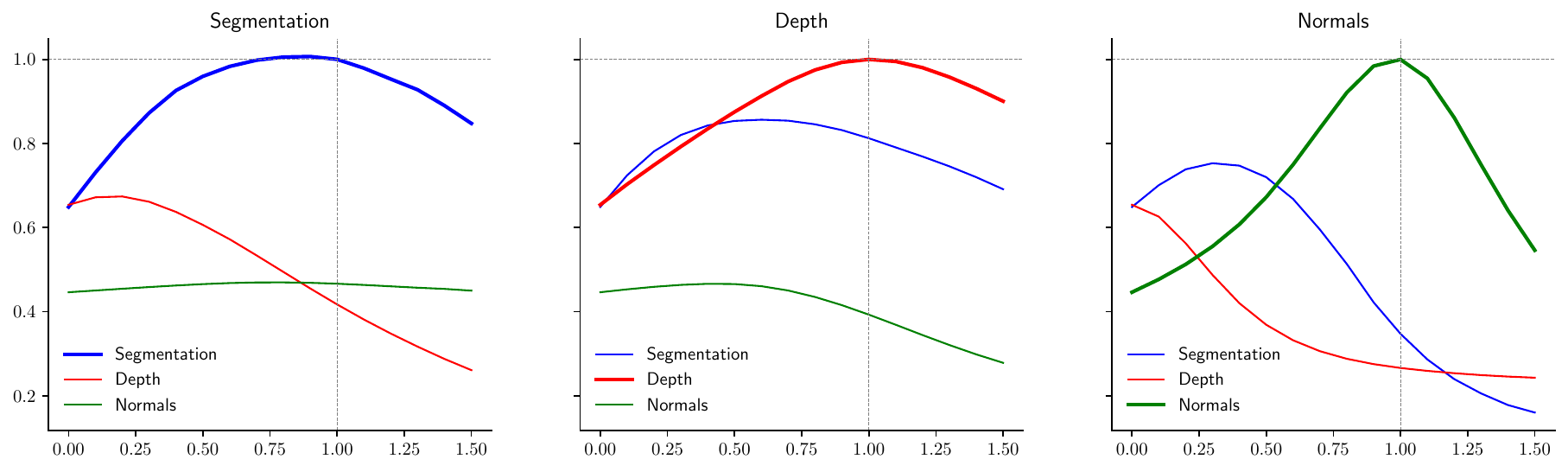}
    \caption{Visualization of task relationships for NYUv2. Each plot presents the normalized performance on each dataset as a function of the task vector added to the model with a coefficient of the x-axis, forming the model $\ptm+\lambda \bm{\tau}$, for $\bm{\tau}\in\{\tv{seg}, \tv{depth},\tv{normals}\}$.}
    \label{fig:nyu_task_relationships}
\end{figure}

\begin{figure}[t]
    \centering
    \includegraphics[width=\linewidth]{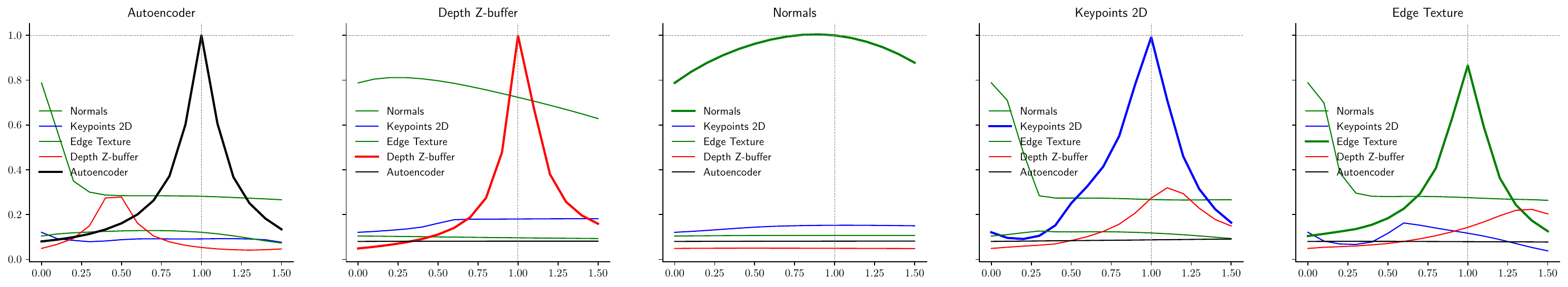}
    \caption{Visualization of task relationships for Taskonomy. Each plot presents the normalized performance on each dataset as a function of the task vector added to the model with a coefficient of the x-axis, forming the model $\ptm+\lambda \bm{\tau}$, for $\bm{\tau}\in\{\tv{auto}, \tv{depth}, \tv{normals}, \tv{keypoints}, \tv{edges}\}$.}
    \label{fig:taskonomy_task_relationships}
\end{figure}

We follow to the methodology outlined in Section~\ref{sec:task_relation} and present visualizations of task relationships for the NYUv2 and Taskonomy datasets. Figure~\ref{fig:nyu_task_relationships} exhibits a consistent pattern with the findings in Section~\ref{sec:task_relation}, whereas Figure~\ref{fig:taskonomy_task_relationships} reveals that Taskonomy tasks lack overlapping feature spaces and are highly sensitive to perturbations in feature representations. This explains the poor performance of naive model merging techniques, which exhibit a degradation of approximately 80\%.

